# Question Embeddings Based on Shannon Entropy
*Solving intent classification task in goal-oriented dialogue system*

Aleksandr Perevalov[1], Daniil Kurushin[1], Rustam Faizrakhmanov[1] and Farida Khabibrakhmanova[2]

[1]*Informational Technologies and Automatic Systems Department, Perm National Research Polytechnic University,*
*9 Pozdeeva st., Perm, Russia*

[2]*Foreign Languages, Linguistics and Translation Department, Perm National Research Polytechnic University,*
*29 Komsomolsky prospect st., Perm, Russia*

*perevalovproduction@gmail.com, dan973@yandex.ru, fayzrakhmanov@gmail.com, farida@pstu.ru*

Abstract: Question-answering systems and voice assistants are becoming major part of client service departments of many organizations, helping them to reduce the labor costs of staff. In many such systems, there is always natural language understanding module that solves intent classification task. This task is complicated because of its case-dependency – every subject area has its own semantic kernel. The state of art approaches for intent classification are different machine learning and deep learning methods that use text vector representations as input. The basic vector representation models such as Bag of words and TF-IDF generate sparse matrixes, which are becoming very big as the amount of input data grows. Modern methods such as word2vec and FastText use neural networks to evaluate word embeddings with fixed dimension size. As we are developing a question-answering system for students and enrollees of the Perm National Research Polytechnic University, we have faced the problem of user's intent detection. The subject area of our system is very specific, that is why there is a lack of training data. This aspect makes intent classification task more challenging for using state of the art deep learning methods. In this paper, we propose an approach of the questions embeddings representation based on calculation of Shannon entropy. The goal of the approach is to produce low dimensional question vectors as neural approaches do and to outperform related methods, described above in condition of small dataset. We evaluate and compare our model with existing ones using logistic regression and dataset that contains questions asked by students and enrollees. The data is labeled into six classes. Experimental comparison of proposed approach and other models revealed that proposed model performed better in the given task.

## 1 INTRODUCTION

Developing of domain-specific question-answering system requires solving natural language understanding tasks. One of them is classification of user's intent, which is frequently solved by machine learning methods. Obviously, machine learning models cannot work with a text itself, and it is required to represent the text as a vector. However, there are some questions about how to represent a text as a vector including the fact that the vector has to represent semantic meaning of the text. Classical methods, such as Bag of Words and TF-IDF are always good baseline for text vectorization in classification tasks, but these methods are producing sparse vectors, that are becoming very big as the data grows.

Modern natural language processing science includes a lot of text-to-vector representations. The major part of them is based on distributive hypothesis: Words that occur in the same contexts tend to have similar meanings [1]. Methods of word to vector representations, such as word2vec [2], FastText [3], Vector Space Model [4], etc. are actually formalization of distributed hypothesis, and are called word embeddings. Therefore, sentence to vector representations — sent2vec [5] and document to vector representations — doc2vec [6] are based on methods mentioned before. All the methods above use neural networks to maximize conditional probability between similar words, that is why they are performing well only when there is enough data for training. However, there are also some cases when researchers or developers come across a lack of data, so modern methods do not work well. Intent Classification on a small dataset is a challenging task for data-hungry state-of-the-art Deep Learning based systems [7].

To summarize previous paragraph, we need to develop a simple, non-data-hungry method of word to dense vector representations which can outperform both classical and modern methods in condition of small and specific dataset. As a solution, we propose the approach of question embeddings based on Shannon entropy calculation, which main idea is to represent word by its' entropy distribution within the questions in given dataset.

The approach will be tested in context of intent classification task within question-answering system for consultation of university students and enrollees. The dataset contains 1300 questions labeled into six classes.

## 2 RELATED WORK

### 2.1 TF-IDF

TF-IDF (term frequency-inverse document frequency) is a classical statistic that reflects how important a word is to a document in a given corpus. Given a document collection $D$, a word $w$, and an individual document $d \in D$, we calculate (1):

$$w_d = f_{w,d} * \log\left(\frac{|D|}{f_{w,d}}\right), \quad (1)$$

where $f_{w,d}$ equals the number of times $w$ appears in $d$, $|D|$ is the size of the corpus, and $f_{w,D}$ equals the number of documents in which $w$ appears in $D$ [8]. In this case, document vector represented as a set of TF-IDF statistics for every word in given collection of documents.

### 2.2 Vector space model

An alternative to TF-IDF is Pointwise Mutual Information (PMI) which is being calculated in vector space model. Let $F$ be a word context (word co-occurrence within window $h$) matrix. Based on context matrix $F$ we calculate matrix $X$ (2), (3) [4]

$$pmi_{i,j} = \log\left(\frac{p_{i,j}}{p_{i*}p_{j*}}\right) \quad (2)$$

$$X = [x_{i,j}], x_{i,j} = \begin{cases} pmi_{i,j}, pmi_{i,j} > 0 \\ 0, pmi_{i,j} \leq 0 \end{cases} \quad (3)$$

In general, $X$ is very sparse that is why truncated singular value decomposition (SVD) is applied (4):

$$\dot{X} = U_k \Sigma_k V_k^T, k < r, \quad (4)$$

where U and Σ are orthonormal matrixes and V is diagonal [9], $r$ is rank of $X$, k is new rank. Given matrix $\dot{X}$ best approximates the original matrix $X$ and minimizes the dimension size. Thus, in matrix $\dot{X}$ i-th row represents a vector of i-th word.

### 2.3 Word2vec and FastText

Word2vec is technique that can be used for learning high-quality word vectors from huge data sets with billions of words, with low dimensionality of word vectors [2]. It has two architectures: Continuous Bag of Words – predicts the current word based on the context, and the Skip-Gram model predicts surrounding words given the current word [2]. These architectures are shown on Figure 1.

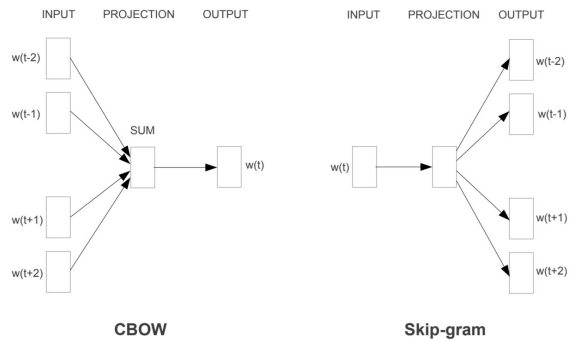

Figure 1: Word2vec architectures.

FastText is an approach proposed by word2vec creators based on Skip-Gram model, where each word is represented as a bag of character n-grams. Let's take word *hello* with n = 3 as an example, it will be represented by the character n-grams:

```
<he, hel, elo, llo, lo>
```

In this case, a word vector is represented as a sum of the vector representations of its n-grams (5):

$$w_t = \sum_{g \in G_w} z_g, \quad (5)$$

where g is an character n-gram, $G_w$ is a set of n-grams appearing in $w$ and $z_g$ is a vector representation of given n-gram.

## 3 METHODS

### 3.1 Data collection

Domain-specificness of the intent classification task requires using relevant data – real-life questions that were asked by students and enrolees. Analysis of existing datasets revealed that there is no open source information required for solving our problem. This is due the fact that the solving classification task is highly specific.

Data collection was performed by scraping open data sources, specifically the following websites: pstu.ru, vk.com/politehperm, abiturient.ru. As a result, 7300 questions were collected. Based on this data, question taxonomy has been developed. The taxonomy consists of six classes: *DOC* – questions about documents, *ENTER* – questions about enrolment process, *ORG* – common questions, *PRIV* – questions about privileges during enrolment, *RANG* – questions related to passing score/exam results, *HOST* – questions about student hostels. Taxonomy is shown on the Figure 2.

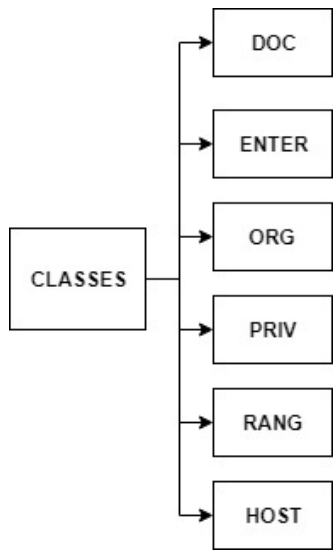

Figure 2: Questions taxonomy.

The example of the training dataset is shown in the table 1. Original data is in Russian, translation is given in parentheses.

Table 1: Training data example.

| Question | Class |
|---|---|
| Можно ли подать документы в субботу? (Is it possible to submit documents on Saturday?) | DOC |
| Когда день открытых дверей? (When is open doors day?) | ORG |
| Когда публикуются списки зачисленных? (When lists of enrolled will be posted?) | RANG |
| Какие документы нужны для заселения в общежитие? (What documents are needed for checking in to students hostel?) | HOST |

Data labelling was made using ipyannotate tool (https://github.com/natasha/ipyannotate) manually. As the result, 1300 questions were labelled. Every question was referred to one class from the taxonomy. Question-class distribution is shown on Figure 3.

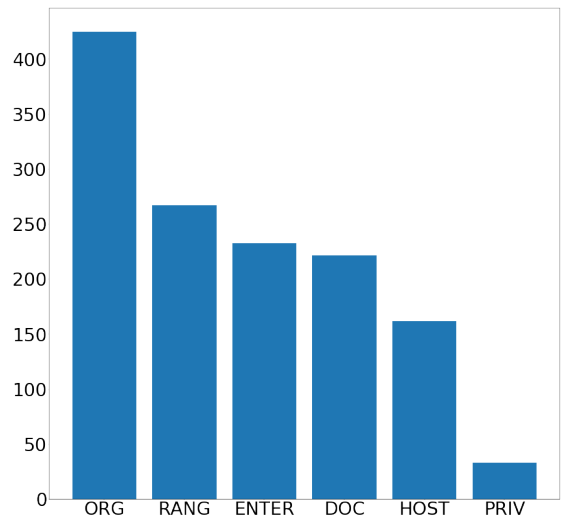

Figure 3: Question-class distribution.

The figure is illustrating imbalance of the dataset: the *ORG* class has much more samples than the average, whereas the *PRIV* class has much less than the average. This aspect has been taken into consideration during classifier evaluation.

### 3.2 Data Preprocessing

Data preprocessing starts with named entity extraction and its transformation to the normal form. It is needed to remove homonyms – words that have the same meaning but different spelling. For named entity extraction, rule-based yargy library (https://github.com/natasha/yargy) was used. Next step in preprocessing is tokenization. After that, every token is being checked for multiple rules, for example if the token is one of the stop-words, or if it contains any Latin letters (system works with Russian language). If one of the rules returns "true" value – then the token is removed, otherwise it remains in the question. All tokens in preprocessed questions are separated by the space character. The code of preprocessing functions is shown in Listing 1.

Listing 1: Preprocessing functions implemented in Python.

```
def preprocess_word(word):
    return stemmer.stem(morph.parse(word)
[0].normal_form.lower())

def preprocess_list(list_):
    new_list = []
    for l in list_:

      for rule in list(ner.rules.keys()):
        parser=ner.Parser(ner.rules[rule])
          for match in parser.findall(l):
            for _ in match.tokens:
              l=l.replace(_.value,rule)

      words = tokenizer.tokenize(l)
      new_words = [preprocess_word(word) for
word in words
      if morph.parse(word)[0].normal_form
not in stopwords and not any(char.isdigit()
for char in word) and not
bool(re.search(r'[a-zA-Z]', word)) and
morph.parse(word)[0].normal_form.lower() not
in custom_stopwords]
      new_list.append(' '.join(w for w in
new_words))
    return new_list
```

Because of word2vec interpretability, preprocessed questions were represented as word2vec embeddings [2] and visualized for cluster analysis. The visualization is shown on the Figure 4.

Visualization showed that some of the classes are clearly separated one from another: *RANG* (marked red on the figure), *DOC* (marked yellow on the figure). Such classes as *ORG* (marked violet on the figure) has many intersections with other classes. This fact undoubtedly has negative impact on effectiveness of the classifier and will be resolved in the future work by redesigning questions taxonomy.

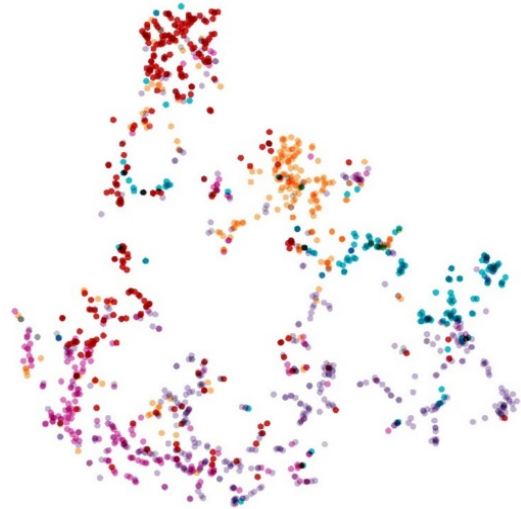

Figure 4: 2D visualization of word2vec embeddings.

### 3.3 Question Embeddings

We propose the approach to the document embeddings or a questions embeddings based on Shannon entropy calculation [10] for every word in the question. One of the Shannon entropy interpretations is a measure of information rate. In this way, the measure of information amount in the word in question is calculated.

First of all, the list of words that appear in the document, is made up. After that, Shannon entropy for every word in the list within every question is calculated (6).

$$e_{ij} = \begin{cases} -p_{ij} \log_2(p_{ij}), p_{ij} = w_{ij} / n_{ij}, w_{ij} > 0 \\ -0.0001, w_{ij} = 0 \end{cases}, \quad (6)$$

where $w_{ij}$ — number of occurrences of *j-th* word in *i-ih* question, $n_{ij}$ — number of words in *i-th* question.

Speaking in terms of machine learning, we calculate matrix where rows (or samples) represent questions, and columns (or features) represent words, so the obtained matrix has 1300 rows (questions) and 1212 columns (features or unique words in training set). Thus, the question is represented by a words entropy vectors. In this case, we need to transpose the matrix, so the rows will represent words, and the columns will represent questions or features (7). In this way, a word, or more precisely a word meaning, is represented by its distribution within the questions in given dataset.

$$M^T = \begin{vmatrix} e_{11} & e_{21} & ... & e_{m1} \\ e_{11} & e_{11} & ... & e_{m1} \\ ... & ... & ... & ... \\ e_{1n} & e_{1n} & ... & e_{m,n} \end{vmatrix} \quad (7)$$

The obtained transposed matrix is sparse (1300 features), thus a dimension reduction using truncated singular value decomposition (4) will be done. In our case, the dimension of the vector will be 200. It is possible by taking first 200 components of decomposed matrixes.

In order to represent a question as a vector, the vectors of words that appear in the question, are to be chosen, and the average of these word vectors, are to be taken (8). The obtained matrix of question vectors and question class will be used as the training set for the classifier.

$$Q_i = \frac{\sum w_j}{count(W)}, w_j \in W \quad (8)$$

where $Q_i$ – vector of $i$-th question, $w_j$ – word vector, $W_i$ – set of the words, that appear in $i$-th question.

## 4 EXPERIMENTS

For experimental testing of the approach proposed, the linear classifier e.g. logistic regression is used. As the dataset has multiple classes, the one vs rest classification method is used. The final model inspired by [11] is shown on the Figure 5.

The proposed Shannon entropy embeddings have been compared with TF-IDF, word2vec and FastText models. Word2Vec and Fast Text word embeddings were transformed to question embeddings by taking the average vector of words' vectors contained in question. During the experiments, dataset was shuffled and split into 5 folds for cross-validation. The evaluation metric for classifier is F1-score (9).

$$F1 = 2 \times \frac{Precision \times Recall}{Precision + Recall} \quad (9)$$

The classification algorithm is logistic regression and classification scheme is "One vs Rest". The results of the experiments are presented in Table 2.

As it can be seen, the proposed method has performed better than existing ones on students and enrollees questions dataset. Its F1-score is 2% higher than the best of the others — TF-IDF. FastText showed the worst result —63% (11% lower than the proposed method). It can be explained by the lack of data in the training set.

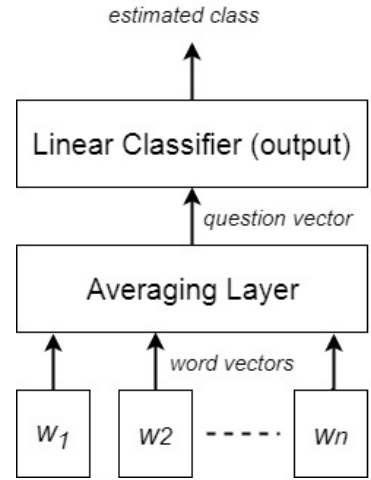

Figure 5: Classification model for the experiments.

Table 2: Classification report – questions dataset.

|  | F1-score | | | |
| --- | --- | --- | --- | --- |
|  | TF-IDF | Word2vec | FastText | **Shannon Entropy** |
| DOC | 0.77 | 0.76 | 0.66 | **0.76** |
| ENTER | 0.62 | 0.59 | 0.58 | **0.65** |
| ORG | 0.73 | 0.66 | 0.65 | **0.74** |
| PRIV | 0.24 | 0.25 | 0.31 | **0.32** |
| RANG | 0.74 | 0.65 | 0.65 | **0.75** |
| HOST | 0.91 | 0.85 | 0.75 | **0.91** |
| Average | 0.72 | 0.67 | 0.63 | **0.74** |

Also, analysis of the results revealed that PRIV class is hardly recognized. This can be explained by the lack of the objects in this class compared with the other classes. To avoid this problem, in future work this class could be merged with other ones.

To make sure in model performance, the proposed approach has been tested on imdb.com reviews dataset. The dataset contains two classes: positive and negative review. Every class contains 10000 reviews written in English. The results of experiments are shown in Table 3.

Table 3: Classification report – IMDB dataset.

|  | F1-score | | | |
| --- | --- | --- | --- | --- |
|  | TF-IDF | Word2vec | Fast Text | **Shannon Entropy** |
| POSITIVE | 0.9 | 0.88 | 0.86 | **0.9** |
| NEGATIVE | 0.9 | 0.88 | 0.86 | **0.9** |
| Average | 0.9 | 0.88 | 0.86 | **0.9** |

It can be seen, that TF-IDF and Shannon entropy showed the same result on F1-score, however, there is difference between dimension sizes in these models: TF-IDF has dimension size equals to 8623 whilst the proposed Shannon entropy model has only 200 (because of applying truncated singular value decomposition, described in 2.2 chapter).

Considering this fact, it can be said that proposed model can store the same information amount with lower dimension size, which can help in improving speed during the data processing.

## 5 CONCLUSIONS

In this work, the approach of question vector representation based on Shannon entropy, has been proposed. For experimental testing, the intent classification task has been suggested. The task was set in terms of voice assistant system for students and enrollees of the university.

The dataset containing students' and enrollees' questions was collected. After that, the taxonomy of the data was designed; the dataset was labeled by classes according to the taxonomy. The approach of question vector representation was designed, implemented and tested.

As the result, the proposed method performed better comparing to the TF-IDF (F1-score is 2% higher), Word2vec (F1-score is 7% higher) and FastText (F1-score is 11% higher).

There was also one experiment on imdb.com reviews dataset that have proved proposed model performance: TF-IDF and Shannon Entropy showed the same result on F-score – 90%, however Shannon Entropy has lower dimension size rather than TF-IDF. This fact can help in improving speed during the data processing without any information loss.

In future work, the redesign of the existing taxonomy for imbalance reduction is planned. Also, modernization of the approach using weighted averaging is going to be done.

The obtained classifier model and the dataset will be used in voice assistant system for students and enrollees consultation. All the data, source code and models described above are available online: https://github.com/Perevalov/intent_classifier.